\documentclass[11pt]{article}

\usepackage{amsmath}
\usepackage{listings}
\usepackage[preprint]{acl}

\usepackage{times}
\usepackage{latexsym}

\usepackage[T1]{fontenc}

\usepackage[utf8]{inputenc}

\usepackage{microtype}

\usepackage{inconsolata}

\usepackage{graphicx}

\title{FASTRIC: Prompt Specification Language for Verifiable LLM Interactions}

\author{Wen-Long Jin \\
  Department of Civil and Environmental Engineering \\
  California Institute for Telecommunications and Information Technology \\
  Institute of Transportation Studies \\
  University of California, Irvine, CA 92697-3600 \\
  \texttt{wjin@uci.edu}}

\begin{document}
\maketitle
\begin{abstract}
Large Language Models (LLMs) can execute complex multi-turn interaction protocols through intelligent interpretation, but lack formal specifications to verify execution against designer’s interaction intent. We introduce FASTRIC, a Prompt Specification Language that makes implicit Finite State Machines (FSMs) explicit in natural language prompts, enabling conformance verification through execution trace analysis. The LLM serves as intelligent execution agent: interpreting designer-encoded FSMs to execute specified behavioral roles in interactions with users and tools. Unlike symbolic specification languages requiring parsers and compilers, FASTRIC leverages LLMs as unified infrastructure—simultaneously parser, interpreter, runtime environment, and development assistant.

FASTRIC guides designers to articulate seven FSM elements (Final States, Agents, States, Triggers, Roles, Initial State, Constraints) that structure multi-turn interactions. Specification formality—ranging from implicit descriptions that frontier models infer to explicit step-by-step instructions for weaker models—serves as a design parameter. We introduce procedural conformance as a verification metric measuring execution adherence to these FSM specifications.

Testing a 3-state kindergarten tutoring FSM across four formality levels and three model scales (14.7B, 685B, 1T+ parameters) reveals that optimal specification formality is a function of model capacity. DeepSeek-V3.2 (685B) achieves perfect conformance (1.00) at L2-L4; ChatGPT-5 (~1T) peaks at L3 (0.90) before collapsing at L4 (0.39); Phi4 (14.7B) shows no stable optimum with high variance (SD=0.16-0.36). These findings reveal model-specific formality ranges—"Goldilocks zones"—where specifications provide sufficient structure without over-constraint, establishing Prompt Specification Engineering as a discipline for creating verifiable interaction protocols for multi-agent systems, transforming multi-turn interaction design from heuristic art to systematic engineering with measurable procedural guarantees.

\end{abstract}

{\bf Keywords}: Prompt Specification Language; Large Language Models; Interaction Protocols; Multi-Agent Systems; Finite State Machines; Specification Formality; Procedural Conformance.

\section{Introduction}

\subsection{Large Language Models: The unprecedented ability and the lack of formal verification}

Large Language Models (LLMs) have transformed human-computer interaction through their unprecedented ability to understand and generate natural language across diverse contexts, enabled by training on massive text corpora. Unlike traditional computational systems requiring separate specification languages, parsers, and runtime environments, LLMs unify these roles through instruction-following in natural language \citep{brown2020language, ouyang2022training}. Beyond simple question-answering, LLMs can execute complex multi-turn workflows across education, research, healthcare, customer service, and collaborative decision-making, interacting with humans, other AI systems, and external tools.

However, current evaluation methodologies for multi-turn, multi-agent interactions (where each turn represents one agent's action) focus primarily on output quality (assessing response accuracy, coherence, or task completion rates) but provide no formal mechanism to verify whether an LLM follows the intended interaction structure specified by prompt designers. This gap between heuristic prompt engineering and formal verification creates serious practical consequences: workflow designers discover prompt failures only through costly real-world deployment, multi-turn behaviors cannot be systematically debugged or refined, and safety-critical applications lack guarantees about behavioral compliance. The absence of formal specification languages for LLM-executable interaction protocols prevents the emergence of systematic engineering practices, leaving multi-turn interaction design as an art rather than a science.

\subsection{FASTRIC: A Prompt Specification Language to Bridge Power and Verifiability}

To bridge this gap, we introduce FASTRIC: a Prompt Specification Language (PSL) designed to systematically structure multi-turn prompts. Unlike simple question-answering exchanges, sustained multi-turn interactions require procedural guarantees across complex, stateful workflows. FASTRIC's core principle is to make the implicit Finite State Machine (FSM) underlying any multi-turn interaction explicit within the prompt itself. This enables designers to inject both procedural structure (how interactions unfold) and domain expertise (what content/constraints apply) into the specification, balancing trade-offs among safety (procedural conformance), efficiency (token usage), transparency (formality level), and quality (task-specific personalization). Unlike approaches that modify model parameters through training or fine-tuning, FASTRIC operates at the specification level—treating prompts as executable behavioral specifications that leverage LLMs' existing instruction-following capabilities.

FASTRIC's architectural commitment positions the LLM as intelligent execution agent: it interprets designer-encoded FSMs to execute specified roles within interaction procedures. Three agent classes participate: the designer (who creates the FSM specification), the LLM (who executes agent roles per FSM—tutor, assistant, evaluator), and users/tools (who interact according to protocol). This bounded intelligent execution distinguishes FASTRIC from approaches where LLMs autonomously generate interaction patterns: here, the LLM enacts designer procedural intent within explicit FSM constraints rather than inventing coordination logic.

A key insight is that optimal specification formality varies across models.\footnote{Appendix~\ref{app:terminology} provides terminology distinguishing FSMs, interaction protocols, interaction procedures, and formality levels (abbreviated as `formality' where context is clear) used throughout this work.} A model with strong instruction-following might correctly execute a high-level description—"ask question, receive answer, provide feedback"—whereas another model (regardless of overall capability) may require detailed procedural steps for reliable execution. FASTRIC accommodates this through varying formality levels, enabling designers to calibrate specification detail to each model's particular strengths. This work focuses on establishing the relationship between formality level and procedural conformance, providing a foundational design principle for the broader specification design space.

FASTRIC exploits a critical property of LLMs: they collapse the traditional language stack. Where conventional systems separate specification languages, parsers, compilers, and runtime environments, the same LLM that executes FASTRIC prompts can also parse them, assist in writing them, and validate conformance through execution traces. This unified infrastructure enables designers to leverage LLMs throughout the specification lifecycle—from generation to verification.

\subsection{Our Contributions}

This paper makes the following key contributions:
\begin{enumerate}
\item A Prompt Specification Language Structured Around FSM Principles: We introduce FASTRIC, a Prompt Specification Language whose core elements (Final States, Agents, States, Triggers, Roles, Initial State, and Constraints) guide designers to explicitly articulate interaction protocols as FSM specifications, transforming informal instructions into formal, verifiable behavioral specifications. 
\item A Prompt Design Methodology Using Formality as a Control Variable: We establish a methodology where specification formality—the level of explicit detail used to describe the FSM—serves as a design parameter, ranging from implicit natural language structures to explicit procedural instructions. 
\item A Prompt Verification Metric Based on Procedural Conformance: We introduce procedural conformance as a metric that quantifies procedural reliability by comparing an agent's execution trace against the canonical FSM specification over test cases, enabling objective evaluation distinct from output quality assessment.
\item An Empirical Design Principle for Formality Selection: We demonstrate that optimal specification formality varies systematically across models. Through experiments across multiple model families and reasoning modes, we identify model-specific formality ranges that balance procedural reliability with specification flexibility, establishing formality calibration as a principled design practice. 
\end{enumerate}

The remainder of this paper is organized as follows. Section~\ref{sec:foundations} reviews the theoretical foundations across three research traditions—formal specification languages, multi-agent interaction protocol engineering, and natural language programming—establishing how their convergence enables FASTRIC. Section~\ref{sec:FASTRIC} presents the FASTRIC Specification Language, detailing its core elements, formality levels, and design methodology. Section~\ref{sec:empirical-validation} provides empirical validation of the formality-conformance relationship across multiple model families and scales. Section~\ref{sec:conclusion} summarizes our contributions and positions FASTRIC as a universal Interaction Protocol Language. Section~\ref{sec:limitations} discusses scope constraints and methodological limitations. Section~\ref{sec:future-directions} outlines research directions for compositional protocols, automated verification, and high-stakes deployment.

\section{Theoretical Foundations}\label{sec:foundations}

FASTRIC synthesizes three independent research traditions: formal verification methods based on Finite State Machines (\S\ref{sec:fsm}), multi-agent protocol engineering including dialogue systems (\S\ref{sec:protocols}), and natural language programming (\S\ref{sec:nlp}). FSM-based formal methods provide the verification framework enabling conformance measurement through trace analysis. Multi-agent protocol engineering provides interaction protocols that map to FSM specifications. Natural language programming—from early decomposition principles to modern LLM instruction-following—provides the execution framework where procedural granularity matches interpreter capacity. Their convergence through LLMs (\S\ref{sec:llms}) enables FASTRIC's contribution: natural language specification with systematic verification.

\subsection{Formal Specification Languages and the FSM Foundation}\label{sec:fsm}
Software engineering separates system specification from implementation, enabling rigorous verification before costly deployment. Formal methods established this practice through mathematically precise specification languages that prove system correctness.

Early foundational work includes Hoare's axiomatic basis for program correctness \citep{hoare1969axiomatic} and Communicating Sequential Processes (CSP) \citep{hoare1978communicating}, which provides algebraic notation for specifying concurrent system interactions. Model checking, independently developed by \citet{clarke1981design} and \citet{queille1982specification}, introduced automated verification by representing systems as FSMs—enabling exhaustive exploration of behaviors within bounded state spaces. Lamport's Temporal Logic of Actions (TLA) \citep{lamport1994temporal} extended these approaches with temporal operators for modeling concurrent systems.

Parallel to formal methods, Domain-Specific Languages (DSLs) emerged to manage complexity through domain-tailored abstractions \citep{vandeursen2000domain}. While DSLs prioritize expressiveness for specific domains, formal specification languages focus on verifiability through mathematical rigor.

These verification approaches, whether symbolic or domain-specific, share a common limitation: they assess declarative correctness—whether outputs match specifications—but cannot verify procedural conformance in multi-turn interactions.

\subsection{Multi-Agent Protocol Engineering and Dialogue Systems}\label{sec:protocols}
Multi-agent systems require interaction protocols specifying interaction patterns: who acts when, what triggers transitions, and how agents respond to each other's actions. The following three approaches addressed protocol specification, each facing distinct limitations.

\begin{enumerate}
\item Agent Communication Languages. The Foundation for Intelligent Physical Agents (FIPA) developed standardized communication protocols drawing on speech act theory. FIPA ACL specified actions (inform, request, propose) with mental state semantics \citep{fipa2002acl}. KQML (Knowledge Query and Manipulation Language) similarly used speech act primitives for agent messaging. Both required symbolic message formats—XML for FIPA, S-expressions for KQML—and explicit protocol parsers.
\item Protocol Specification Languages. Formal methods developed protocol specification through process algebras and choreography languages. CSP provided algebraic notation for concurrent system interactions, enabling formal reasoning about deadlock and liveness properties. FIPA defined interaction protocols—Contract Net, Request Protocol—as sequence diagrams describing expected message exchanges. WS-CDL and BPEL specified service choreographies for web services \citep{peltz2003web}. These approaches provided rigorous specifications but remained declarative: they described protocols for verification rather than providing executable specifications that agents could directly interpret.
\item Multi-Turn Dialogue Systems. Task-oriented dialogue (TOD) systems employ state machines for dialogue management to achieve specific goals like booking or customer service. These state machines are architectural implementations rather than verifiable specifications—they manage conversation flow internally but lack formal specification languages for protocol design or conformance verification. Research focuses on context management, intent recognition, and dialogue state tracking rather than procedural guarantees.
\end{enumerate}

These approaches shared a common limitation: separating specification from execution. Agent communication languages and protocol specification languages required symbolic representations, while dialogue systems implemented state machines architecturally without formal specification languages.

\subsection{Natural Language Programming and Instruction-Following}\label{sec:nlp}

The ability to instruct computational systems using natural language has long been pursued as an alternative to formal specification languages. This pursuit faced a fundamental challenge: natural language's flexibility appeared incompatible with the precision required for reliable program execution.

Dijkstra dismissed natural language programming as too imprecise for computation \citep{dijkstra1978foolishness}, focusing on ambiguity inherent in natural descriptions. Early systems like Attempto Controlled English \citep{fuchs1996attempto} restricted English to machine-parsable subsets but sacrificed naturalness and failed to scale.

\citet{ballard1979programming} recognized that breaking complex goals into finer-grained procedural steps could bridge the gap between human intent and machine execution. Their core insight—procedural granularity affects execution reliability—remained theoretical for four decades without capable natural language understanding systems.

LLMs enabled interpretation of unrestricted natural language through instruction-following behavior. Training through RLHF and instruction-tuning \citep{ouyang2022training} taught models to interpret first-person instructions (``I will ask'') as executable commitments. Modern prompting techniques empirically validate this decomposition insight: Chain-of-Thought \citep{wei2022chain} breaks reasoning into explicit steps, while ReAct \citep{yao2022react} interleaves reasoning with tool use.

However, these techniques improve output quality heuristically but provide no conformance verification framework for multi-turn prompts. Multi-IF \citep{he2024multi} reveals performance degradation across turns (frontier models drop from 87.7\% to 70.7\% accuracy turn 1 to 3), but focuses on instruction accumulation rather than procedural protocols with explicit state machines.

\subsection{Convergence: Why Natural Language Specification Became Possible}\label{sec:llms}
The aforementioned three traditions developed independently for decades, each accepting fundamental constraints as insurmountable: formal methods required symbolic languages for verification algorithms, protocol engineering required explicit parsers without natural language understanding, and natural language programming remained aspirational without reliable interpretation.

LLMs dissolve these barriers through instruction-following behavior, treating natural language specifications as executable protocols \citep{brown2020language, ouyang2022training}. This capability enables three convergent possibilities. First, LLMs execute natural language instructions directly, enabling conformance measurement through execution trace analysis rather than requiring formal proof systems. Second, domain experts can describe interaction patterns in natural language rather than learning specialized notations like process algebras or sequence diagrams. Third, LLMs collapse the traditional language stack—the same system executes instructions, parses them, assists in writing them, and validates conformance through execution traces.

However, this convergence reveals a critical gap: while LLMs can execute natural language prompts, no systematic framework exists for specifying and verifying multi-turn prompts. Instruction-following benchmarks measure output quality, dialogue systems focus on application-level goals, and prompting techniques remain heuristic. The field lacks specification languages mapping natural language to verifiable FSM structures, formality calibration principles, and conformance metrics enabling systematic debugging.

\section{The FASTRIC Specification Language}\label{sec:FASTRIC}
\subsection{Design Foundations}
FASTRIC instantiates the convergence identified in \S\ref{sec:llms} through three architectural commitments that enable specification-level interaction protocol design. The LLM serves as intelligent execution substrate: interpreting designer-encoded FSMs to execute specified behavioral roles in interactions with users and tools. Unlike approaches that modify model parameters through training or fine-tuning, FASTRIC treats prompts as executable behavioral specifications, leveraging LLMs' existing instruction-following capabilities.

Specification design requires navigating four competing objectives: safety (procedural conformance—ensuring agents execute specified interaction patterns without deviation, critical for applications requiring behavioral guarantees), efficiency (token usage—minimizing specification overhead while maintaining sufficient detail), transparency (specification explicitness—enabling human stakeholders to audit interaction logic), and quality (contextual flexibility—preserving model capacity to generate appropriate responses within procedural constraints). FASTRIC's formality levels enable designers to calibrate these trade-offs systematically.

First, FASTRIC makes implicit FSM structure explicit through seven elements (Final States, Agents, States, Triggers, Roles, Initial State, Constraints) that map natural language specifications to formal FSM components $(Q, \Sigma, \delta, q_0, F)$. This enables designers to specify interaction protocols with both procedural structure (how interactions unfold) and domain expertise (what content/constraints apply) while enabling procedural conformance measurement through execution trace analysis rather than requiring symbolic proof systems.

Second, FASTRIC treats specification formality—the degree of structural and behavioral explicitness—as a controllable design variable. Four formality levels (L1-L4) calibrate specification detail to interpreter capacity: frontier models interpret implicit high-level descriptions while weaker models require explicit step-by-step instructions. This accommodates the key insight that optimal specification formality varies across models, enabling designers to match specification detail to each model's instruction-following strengths.

Third, FASTRIC exploits LLMs' collapse of the traditional language stack. The same model that executes FASTRIC specifications can parse them during design, assist in writing them, and validate procedural conformance through execution traces—enabling systematic lifecycle management from generation to verification within a unified computational framework.

\subsection{The Seven Elements and FSM Mapping}\label{sec:seven-elements}

FASTRIC specifies interaction protocols through seven elements mapping natural language to formal FSM components. Table~\ref{tab:fastric-fsm} illustrates these mappings using the kindergarten math tutor protocol.

\begin{table*}[t]
\centering
\caption{FASTRIC Elements and FSM Mapping}
\label{tab:fastric-fsm}
\begin{tabular}{lll}
\hline
\textbf{Element} & \textbf{FSM Component} & \textbf{Kindergarten Tutor Example} \\
\hline
Final States (\textbf{F}) & Terminal states $q \in F$ & Implicit (indefinite tutoring) \\
Agents (\textbf{A}) & Autonomous entities & AI tutor (``I''), student (``you'') \\
States (\textbf{S}) & State space $Q$ & State 0: Init, State 1: EASY, State 2: HARD \\
Triggers (\textbf{T}) & Transition function $\delta$ & ``MORE'' (self-loop), ``CHANGE'' (switch states) \\
Roles (\textbf{R}) & Agent behaviors per state & Ask question $\to$ Evaluate $\to$ Prompt navigation \\
Initial State (\textbf{I}) & Start state $q_0$ & State 0: Choose difficulty \\
Constraints (\textbf{C}) & Global invariants & Never reveal answer unless correcting \\
\hline
\end{tabular}
\end{table*}

\begin{figure}[t]
\centering
\includegraphics[width=\columnwidth]{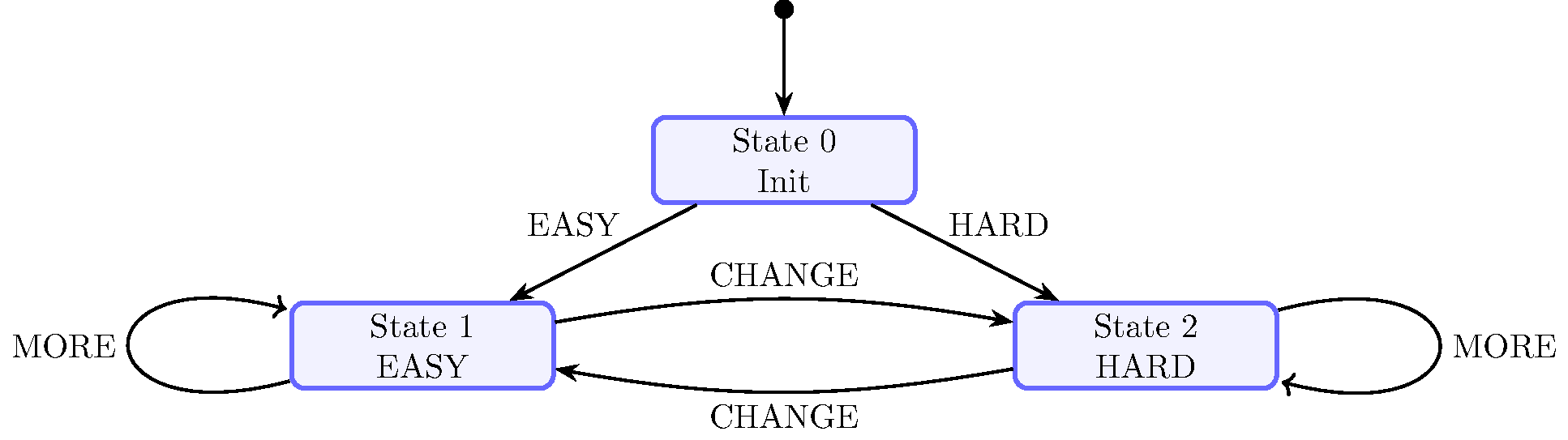}
\caption{Kindergarten math tutor FSM structure. State 0 (Init): student chooses EASY$\to$State 1 or HARD$\to$State 2. States 1--2 execute symmetric loops: ask question, wait for answer, evaluate answer, prompt navigation (MORE loops current state, CHANGE switches states).}
\label{fig:fsm-structure}
\end{figure}

Figure~\ref{fig:fsm-structure} shows the protocol's 3-state machine with symmetric Easy/Hard modes. State 0 branches to State 1 (EASY) or State 2 (HARD) based on student choice. Both states implement identical behavioral loops—ask question, wait for answer, evaluate answer, prompt navigation—with ``MORE'' triggering self-loops and ``CHANGE'' triggering state switches. A FASTRIC specification maps to FSM tuple $(Q, \Sigma, \delta, q_0, F)$ where states $Q$ derive from \textbf{S}, initial state $q_0$ from \textbf{I}, final states $F$ from \textbf{F}, transitions $\delta$ from \textbf{T}, agent behaviors from \textbf{R}, and safety properties from \textbf{C}.

FASTRIC controls procedural conformance through two orthogonal formality dimensions. \textit{Structural formality} determines FSM explicitness: implicit specifications (L1--L2) condense state logic requiring models to infer boundaries, while explicit specifications (L3--L4) separate states into distinct blocks with jump/stay imperatives. \textit{Behavioral formality} determines action granularity: low-formality provides condensed descriptions (``ask ONE question''), while high-formality decomposes actions into numbered sequences with wait statements and imperatives (``(1) I will ask (2) I wait (3) I will evaluate ONLY `Correct!' OR `Wrong'\,'').  Section~\ref{sec:formality-levels} details how L1--L4 combine these dimensions, enabling designers to calibrate specification detail to model capacity.

\subsection{Formality Levels as Control Variable}\label{sec:formality-levels}

FASTRIC's key innovation treats specification formality—the degree of structural and behavioral explicitness—as a controllable design parameter. The same canonical FSM (Figure~\ref{fig:fsm-structure}) can be specified at varying formality levels, enabling designers to calibrate detail to interpreter capacity. We define four canonical levels (Appendix~\ref{app:formality-levels}) spanning from implicit natural language to near-algorithmic instructions. For example, L1's implicit ``I will ask ONE math question based on your choice'' contrasts with L4's explicit ``(1) I will ask ONE easy question (2) I wait for your answer (3) I will evaluate\dots''. L1 condenses both difficulties and all actions; L4 separates difficulty modes (structural) and decomposes actions into numbered steps with waits (behavioral).

The L1$\to$L4 progression makes FSM structure increasingly explicit. L1--L2 unify difficulty modes within abstract descriptions, requiring models to infer state separation. L3--L4 explicitly separate EASY and HARD into distinct structural blocks using jump/stay imperatives—resembling structured GOTO. Within each state, formality controls action granularity: L1 provides condensed descriptions; L2 prescribes output formats (ONLY ``Correct!'' OR ``Wrong, [X]''); L3 sequences actions into numbered sub-steps; L4 adds wait statements and emphasized imperatives (MUST, ONLY), with global invariants separated into Critical Rules sections.

The \textit{Formality-Capacity Principle} formalizes this relationship: for model $M$ and protocol $P$, optimal formality $f^*(M,P)$ balances safety (procedural conformance), efficiency (token usage), transparency (readability), and quality (task flexibility). L1 maximizes efficiency through brevity but sacrifices safety in weaker models; L4 maximizes safety through explicitness but increases token costs and may over-constrain frontier models. Section~\ref{sec:empirical-validation} validates the procedural conformance dimension empirically.

\section{Empirical Validation}\label{sec:empirical-validation}

We measure procedural conformance across three contemporary LLMs executing a standardized tutoring interaction procedure (kindergarten math tutor protocol + 21-turn test sequence) specified at varying FASTRIC formality levels.

\subsection{Experimental Design}
\label{sec:experimental-design}

We implemented a 3-state FSM (State 0: Init, State 1: EASY, State 2: HARD), shown in Figure~\ref{fig:fsm-structure}, with student-controlled navigation via ``MORE''/``CHANGE'' commands. Table~\ref{tab:test-sequence} presents the standardized 21-turn test sequence exercising all state transitions and error handling.

\begin{table*}[t]
\centering
\small
\caption{Standardized Test Sequence (21 Turns)}
\label{tab:test-sequence}
\begin{tabular}{lll}
\hline
\textbf{Turn} & \textbf{Student Input} & \textbf{Expected Tutor Behavior} \\
\hline
1 & & Ask ``Choose EASY or HARD'' \\
2,3 & EASY & Enter easy mode (Ask 1st easy question) \\
4,5 & 5 & Evaluate answer $\to$ Prompt navigation \\
6,7 & more & Loop easy mode (Ask 2nd easy question) \\
8,9 & [Correct Answer to the Tutor's 2nd Easy Question] & Evaluate answer $\to$ Prompt navigation \\
10,11 & change & Switch to hard mode (Ask 1st hard question) \\
12,13 & [Incorrect Answer to the Tutor's 1st Hard Question] & Evaluate answer $\to$ Prompt navigation \\
14,15 & yes & Re-prompt navigation (ambiguous input) \\
16,17 & what & Re-prompt navigation (ambiguous input) \\
18,19 & change & Switch to easy mode (Ask 3rd easy question) \\
20,21 & [Correct Answer to the Tutor's 3rd Easy Question] & Evaluate answer $\to$ Prompt navigation \\
\hline
\multicolumn{3}{l}{\small \textit{Note:} Even turns = student inputs; odd turns = tutor actions.} \\
\end{tabular}
\end{table*}

Four specifications (L1=implicit $\to$ L4=maximally explicit) derived from Section~\ref{sec:formality-levels} appear in Appendix~\ref{app:specifications}. Three models were tested: DeepSeek-V3.2 (685B, no thinking mode), ChatGPT-5 ($\sim$1T, thinking always on), and Phi4-14.7B (14.7B, no thinking mode).\footnote{DeepSeek-V3.2 accessed via \url{https://chat.deepseek.com}, ChatGPT-5 via \url{https://chatgpt.com}, Phi4-14.7B via Ollama (ollama.com) local deployment, all tested November 2025.}

Procedural conformance score equals the number of correctly executed turns (following FSM specification) divided by 21, measured up to the first violation. Student turns (even-numbered) are always correct by design, as inputs follow the standardized script in Table~\ref{tab:test-sequence}. Tutor turns (odd-numbered) are correct if the AI executes the required behavioral loop: ask question, wait for answer, evaluate answer, and prompt navigation. State transitions must follow specified triggers (MORE loops current state, CHANGE switches states). The critical constraint is never revealing answers before student responses. A violation in any tutor turn terminates the interaction.

Twenty independent runs per condition (3 models $\times$ 4 formality levels = 240 runs) were conducted, with each run starting a new chat session to isolate specification effects from conversation history.

\subsection{Results}
\label{sec:results}

Table~\ref{tab:conformance-scores} presents mean procedural conformance scores across conditions. Complete run-level data (all 240 runs) are available in the supplementary materials on OSF: \url{https://doi.org/10.17605/OSF.IO/PV6R3}.

\begin{table*}[t]
\centering
\caption{Procedural Conformance by Model and Formality Level}
\label{tab:conformance-scores}
\begin{tabular}{lccccc}
\hline
\textbf{Model} & \textbf{Parameters} & \textbf{L1} & \textbf{L2} & \textbf{L3} & \textbf{L4} \\
\hline
DeepSeek-V3.2 & 685B & 0.67 (0.16) & 1.00 (0.00) & 1.00 (0.00) & 1.00 (0.00) \\
ChatGPT-5 & $\sim$1T & 0.46 (0.06) & 0.63 (0.23) & 0.90 (0.16) & 0.39 (0.26) \\
Phi4-14.7B & 14.7B & 0.59 (0.16) & 0.32 (0.27) & 0.52 (0.28) & 0.75 (0.36) \\
\hline
\multicolumn{6}{l}{\small \textit{Note:} Values show mean (SD) conformance rate. } \\
\multicolumn{6}{l}{\small Complete run-level data available at \url{https://doi.org/10.17605/OSF.IO/PV6R3}.} \\
\end{tabular}
\end{table*}

\begin{figure}[t]
\centering
\includegraphics[width=\columnwidth]{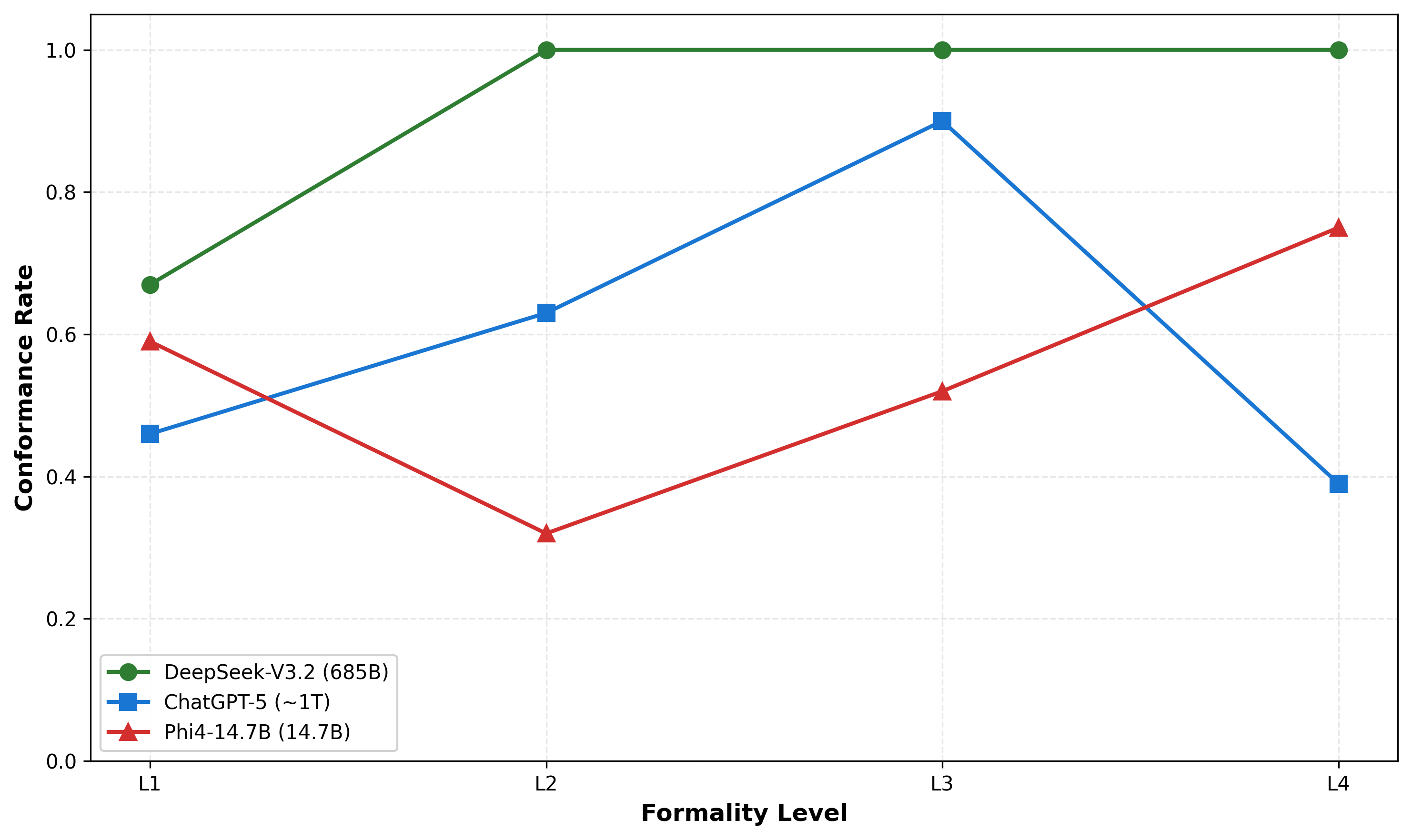}
\caption{Mean procedural conformance by model and formality level}
\label{fig:conformance}
\end{figure}

As shown in Figure~\ref{fig:conformance}, DeepSeek-V3.2 achieved perfect deterministic execution (1.00, SD=0.00) at L2--L4 across all 60 runs, with sharp improvement from L1 (0.67). ChatGPT-5 exhibited an inverted-U pattern peaking at L3 (0.90), then collapsing to L4 (0.39)—a 57\% decline. Phi4-14.7B showed no stable optimum, ranging from 0.32 (L2) to 0.75 (L4) with high variance (SD=0.16--0.36).

\subsection{Discussion}
\label{sec:discussion}

Optimal formality varies by model: DeepSeek performs identically at L2--L4, ChatGPT peaks sharply at L3, and Phi4 lacks clear optimum. Two failure modes occur across models and formality levels. First, confirmation-seeking rather than direct state transitions: when the student inputs ``change'' (turn 10), the tutor asks ``Do you want to switch to HARD?'' instead of immediately switching states. Second, misinterpreting ambiguous inputs: when the student inputs ``yes'' (turn 14), the tutor treats this as ``MORE'' instead of re-prompting with valid options. ChatGPT's L4 collapse exhibits a distinct failure mode: case-sensitive brittleness where the model rejects ``more'' as invalid because the specification prescribes ``MORE''.

Appendix~\ref{app:conformance-distributions} reveals the distribution properties underlying these patterns. DeepSeek-V3.2's zero variance at L2--L4 (horizontal lines in box plots) demonstrates deterministic execution. ChatGPT-5's wide L4 distribution indicates instability rather than consistent poor performance—some runs achieve perfect scores while others fail catastrophically. Phi4-14.7B's overlapping distributions across all formality levels, combined with high variance (SD=0.16--0.36), suggest architectural limitations that specification design cannot overcome.

These findings validate FASTRIC's core design principle: specification formality must be calibrated to model characteristics. The effectiveness of a FASTRIC specification depends not only on FSM correctness but on matching formality level to interpreter capacity.

\section{Conclusion}
\label{sec:conclusion}

FASTRIC introduces a Prompt Specification Language making implicit FSM structure explicit in natural language prompts, enabling procedural conformance measurement. Experiments across three models (DeepSeek-V3.2, ChatGPT-5, Phi4-14.7B) and four formality levels reveal that optimal formality varies by model capacity: DeepSeek achieves perfect execution (1.00) at L2--L4, ChatGPT peaks at L3 (0.90) before collapsing at L4 (0.39), and Phi4 shows no stable optimum—validating model-specific ``Goldilocks zones'' balancing procedural structure against over-constraint.

While current approaches focus on output quality through heuristic prompt engineering, FASTRIC provides the first formal mechanism to verify procedural conformance in multi-turn interactions. This transforms multi-turn interaction design from art—where failures emerge only through costly deployment—into engineering with verifiable guarantees for safety-critical applications.

Beyond LLM applications, FASTRIC's protocol-centric architecture positions it as a universal Interaction Protocol Language (IPL) for heterogeneous agent systems. Designers encode interaction intent through FSM structure while various computational substrates—LLMs, humans, symbolic AI, hybrid systems—decode specifications into conformant behaviors. This addresses the expertise gap \citet{rahwan2019machine} identified between behavioral scientists who cannot implement algorithms and computer scientists who lack experimental methodology training, enabling FASTRIC to scale from today's LLM-executed applications to tomorrow's multi-substrate intelligence platforms.

\section{Limitations}
\label{sec:limitations}

First, we validated a single bounded protocol (kindergarten math tutoring) across three model families. Generalization to diverse interaction types—multi-party negotiations, adaptive workflows, heterogeneous agent topologies—requires further validation.

Second, procedural conformance measurement relied on manual annotation across 240 runs. Scalable deployment requires automated verification, which introduces recursive validation challenges (verifying the verifier) not addressed in this work.

Third, our formality-capacity analysis spans three scales (14.7B, 685B, ~1T parameters) but does not systematically quantify efficiency-quality trade-offs. L4's safety gains may impose token costs or over-constraint penalties beyond our conformance-focused evaluation.

\section{Future Directions}
\label{sec:future-directions}

First, extending FASTRIC to compositional protocols: nested and concurrent FSMs where multiple protocols interact, adaptive protocols that modify structure during execution, and multi-agent topologies with conflicts or adversarial dynamics.

Second, automating conformance measurement through "Verifier Agents"—specialized LLMs monitoring procedural adherence—and developing specification synthesis where LLMs bootstrap their own protocols via simulation-based design.

Third, deploying FASTRIC in high-stakes environments where procedural deviation is safety failure, not error. Autonomous systems from AV-pedestrian negotiation to robotic surgical assistants will use natural language to execute coordinated actions—FASTRIC offers a path toward treating natural language directives as verifiable safety constraints.

\section*{Acknowledgments}

We thank Shiqi Wang and Hongyang Chu (college students, IELTS preparation), Christopher Jin (high school student, English/Chinese/history tutoring), Jooneui Hong (PhD student, English learning), and Wenbin Jin (MS-level professional, mathematics tutoring in Chinese) for testing early FASTRIC versions. Christopher's observation that FASTRIC felt like ``coding with prompts'' provided critical insight into the programming language perspective. Claude (Anthropic) and Gemini (Google) were used for literature search, brainstorming, drafting assistance, and editing; all core concepts, experimental design, analysis, and final content decisions remain the authors' original intellectual contributions. All errors and limitations remain our own.


\clearpage
\appendix

\section{FASTRIC Terminology}
\label{app:terminology}

\begin{center}
\footnotesize
\renewcommand{\arraystretch}{0.9}
\begin{tabular}{p{3cm}p{5cm}p{6cm}}
\hline
\textbf{Term} & \textbf{Definition} & \textbf{Kindergarten Tutor Example} \\
\hline
FSM (Finite State Machine) & Formal model with states ($Q$), alphabet ($\Sigma$), transitions ($\delta$), initial state ($q_0$), final states ($F$) & 3 states: Init, EASY, HARD; transitions via MORE/CHANGE \\
Interaction Protocol & Agent's behavioral specification (FSM defining roles, states, transitions) for responding to inputs & Tutor protocol with EASY/HARD states: ask $\to$ evaluate $\to$ prompt \\
FASTRIC Specification & Natural language protocol description using FASTRIC's seven elements at chosen formality level & The L1--L4 prompts in Appendix~\ref{app:specifications} \\
Interaction Procedure & Complete multi-agent specification comprising protocol(s) and expected patterns & Tutor protocol responding to student commands (MORE/CHANGE) \\
Formality Level & Degree of procedural explicitness (L1=implicit $\to$ L4=maximally explicit) & L1: unified states; L4: separated states, waits, MUST imperatives \\
Agent & Entity with specified role in protocol design and execution & Designer, LLM tutor, student \\
Intelligent Execution & LLM's dual role as semantic interpreter and bounded executor & LLM tutor executes designer-specified protocol per FSM \\
Procedural Conformance & Metric measuring execution adherence; ratio of correct transitions to total & 21/21 = 1.00 (perfect); 10/21 = 0.48 (failed turn 11) \\
State Transition & Movement between FSM states triggered by inputs & ``CHANGE'' $\to$ tutor transitions EASY $\to$ HARD \\
Execution Trace & Sequence of agent actions across turns for conformance measurement & 21-turn sequence: ask $\to$ answer $\to$ evaluate $\to$ \dots \\
\hline
\end{tabular}
\renewcommand{\arraystretch}{1}
\end{center}


\clearpage
\section{Formality Levels for Kindergarten Math Tutor} \label{app:formality-levels}

\begin{center}
\footnotesize
\renewcommand{\arraystretch}{0.9}
\begin{tabular}{p{3cm}p{5cm}p{6cm}}
\hline
\textbf{Level} & \textbf{Key Characteristics} & \textbf{Step 1 Specification (Excerpt)} \\
\hline
L1 (Implicit) & Unified states, condensed actions & ``I will ask ONE math question based on your choice, then ask: `MORE at same level, or CHANGE difficulty?'\,'' \\
L2 (Semi-Explicit) & Visible transitions, prescribed formats & ``Step 1: Ask ONE question based on choice. Evaluate by saying ONLY `Correct!' OR `Wrong, answer is [X]'. Ask: `MORE at same level, or CHANGE?' If MORE, stay; if CHANGE, change difficulty.'' \\
L3 (Explicit) & Separated states, jump logic & ``Step 1: EASY problems. (1) Ask ONE easy question (2) Evaluate ONLY `Correct!' OR `Wrong, [X]' (3) Ask: `MORE at easy, or CHANGE to hard?' (4) If MORE, stay; if CHANGE, jump to Step 2.'' \\
L4 (Maximally Explicit) & Wait statements, imperatives, constraints & ``Step 1: EASY. (1) Ask ONE easy question (2) I wait (3) Evaluate ONLY `Correct!' OR `Wrong, [X]' (4) I MUST ask exactly: `MORE at easy, or CHANGE to hard?' (5) If MORE, stay; if CHANGE, jump to Step 2.'' \\
\hline
\multicolumn{3}{p{15.5cm}}{\small \textit{Note:} Bold indicates added mechanisms. All excerpts show Step 1 (EASY mode) for direct comparison. Complete specifications in Appendix~\ref{app:specifications}.} \\
\end{tabular}
\renewcommand{\arraystretch}{1}
\end{center}
\clearpage

\section{Kindergarten Math Tutor Specifications (L1--L4)}
\label{app:specifications}

\subsection{L1 Specification}
\label{app:spec-l1}

\begin{lstlisting}
===== INSTRUCTION BEGINS =====
Note: "I" refers to the AI tutor of kindergarten math; "you" refers to the kindergarten student.

## Step 0
1. I will ask you to choose between EASY and HARD problems.

## Step 1
1. I will first ask ONE math question based on your choice of difficulty level.
2. After you answer the question, I will then ask you: "MORE at the same level, or CHANGE difficulty level?"
===== INSTRUCTION ENDS =====
\end{lstlisting}

\subsection{L2 Specification}
\label{app:spec-l2}

\begin{lstlisting}
===== INSTRUCTION BEGINS =====
Note: "I" refers to the AI tutor of kindergarten math; "you" refers to the kindergarten student.

## Step 0
I will start with this step.
1. I will ask you to choose between EASY and HARD problems.
2. I will wait for your answer.
3. I will proceed to Step 1.

## Step 1
I will now enter a loop based on your choice.
1. I will ask ONE math question based on your choice of difficulty level.
2. I will evaluate the answer by saying ONLY "Correct!" OR "Wrong, the answer is [X]".
3. After evaluating, I must ask: "MORE at the same level, or CHANGE difficulty level?".
4. If your command is "MORE", I will stay at the same difficulty level; if your command is "CHANGE", I will change the difficulty level.
===== INSTRUCTION ENDS =====
\end{lstlisting}

\subsection{L3 Specification}
\label{app:spec-l3}

\begin{lstlisting}
===== INSTRUCTION BEGINS =====
Note: "I" refers to the AI tutor of kindergarten math; "you" refers to the kindergarten student.

## Step 0
I will start with this step.
1. I will ask you to choose between EASY and HARD problems.
2. If you choose EASY, I will jump to Step 1; if you choose HARD, I will jump to Step 2.

## Step 1: EASY problems
1. I will ask ONE easy math question.
2. I will evaluate the answer by saying ONLY "Correct!" OR "Wrong, the answer is [X]".
3. After evaluating, I must ask: "MORE at the easy level, or CHANGE to the hard level?".
4. If your command is "MORE", I will stay in this step; if your command is "CHANGE", I will jump to Step 2: HARD problems.

## Step 2: HARD problems
1. I will ask ONE hard math question.
2. I will evaluate the answer by saying ONLY "Correct!" OR "Wrong, the answer is [X]".
3. After evaluating, I must ask: "MORE at the hard level, or CHANGE to the easy level?".
4. If your command is "MORE", I will stay in this step; if your command is "CHANGE", I will jump to Step 1: EASY problems.
===== INSTRUCTION ENDS =====
\end{lstlisting}

\subsection{L4 Specification}
\label{app:spec-l4}

\begin{lstlisting}
===== INSTRUCTION BEGINS =====
Note: "I" refers to the AI tutor of kindergarten math; "you" refers to the kindergarten student.

## Step 0
I will start with this step.
1. I will ask you to choose between EASY and HARD problems.
2. I will wait for your answer.
3. If you choose EASY, I will jump to Step 1; if you choose HARD, I will jump to Step 2.

## Step 1: EASY problems
1. I will ask ONE easy math question.
2. I wait for your answer.
3. I will evaluate the answer by saying ONLY "Correct!" OR "Wrong, the answer is [X]".
4. After evaluating, I MUST ask the following question exactly: "MORE at the easy level, or CHANGE to the hard level?"
5. If your command is "MORE", I will stay in this step; if your command is "CHANGE", I will jump to Step 2: HARD problems.

## Step 2: HARD problems
1. I will ask ONE hard math question.
2. I wait for your answer.
3. I will evaluate the answer by saying ONLY "Correct!" OR "Wrong, the answer is [X]".
4. After evaluating, I MUST ask the following question exactly: "MORE at the hard level, or CHANGE to the easy level?"
5. If your command is "MORE", I will stay in this step; if your command is "CHANGE", I will jump to Step 1: EASY problems.

## Critical Rules
1. I must never answer a math problem for you unless I am correcting a wrong answer.
2. I must stick to this workflow exactly. I do not add extra steps or commentary unless specified.
3. If you provide an invalid command (not "MORE" or "CHANGE"), I must re-prompt you with the valid options.
===== INSTRUCTION ENDS =====
\end{lstlisting}

\clearpage

\section{Procedural conformance distributions by model and formality level. }\label{app:conformance-distributions}

\begin{figure}[h]
\centering
\caption{Box plots show median (black line), interquartile range (box), full range (whiskers), and mean (white diamond).}
\includegraphics[width=\columnwidth]{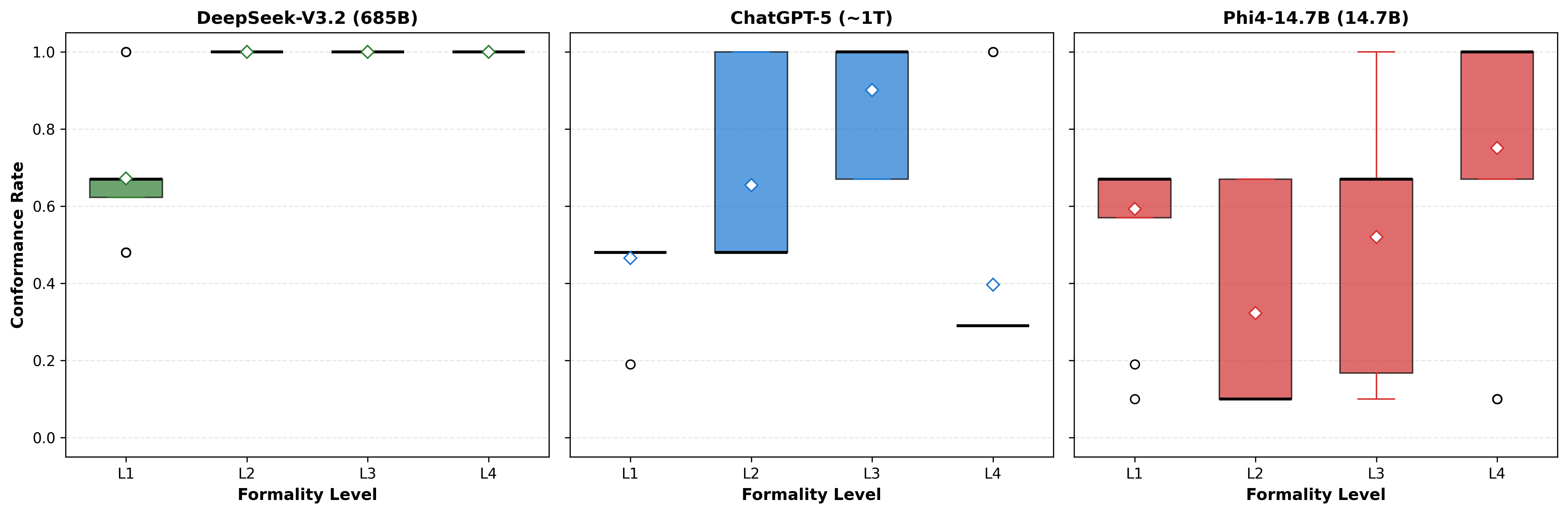}
\end{figure}

\end{document}